%% file: main.tex
\documentclass[conference]{IEEEtran}
\IEEEoverridecommandlockouts
\usepackage{cite}
\usepackage{amsmath,amssymb,amsfonts,amsthm}
\usepackage{algorithmic}
\usepackage{booktabs}
\usepackage{comment}
\usepackage{graphicx}
\usepackage{textcomp}
\usepackage{csquotes}
\usepackage{textcomp}
\usepackage{multirow}
\usepackage[hidelinks]{hyperref}
\usepackage{xcolor}
\usepackage[ruled, linesnumbered, vlined]{algorithm2e}
\usepackage{listings}%
\usepackage{subfig}
\def\BibTeX{{\rm B\kern-.05em{\sc i\kern-.025em b}\kern-.08em
    T\kern-.1667em\lower.7ex\hbox{E}\kern-.125emX}}

\def\equationautorefname~#1\null{Eq.~(#1)\null}

\begin{document}

\input{misc/title_and_authors}

\maketitle

\input{misc/abstract}


\section{Introduction}
\label{sec:introduction}
\input{sections/1_introduction}

\section{Related Work}
\label{sec:related_work}
\input{sections/2_related}

\section{Problem Setup}
\label{sec:problem_setup}
\input{sections/3_problem}

\section{Proposed Solution}
\label{sec:proposed_solution}
\input{sections/4_solution}

\section{Application Domain}
\label{sec:mofs}
\input{sections/5_mof_application}
\section{Experimental Design \& Results}
\label{sec:exp_design_results}
\input{sections/6_results}

\section{Conclusions}
\label{sec:conclusions}
\input{sections/7_conclusions}

\section*{Acknowledgment}
This work was supported by Laboratory Directed Research and Development (LDRD) funding from Argonne National Laboratory, provided by the Director, Office of Science, Office of Basic Energy Science, Division of Chemical Sciences, Geosciences, and Biosciences of the U.S. Department of Energy under Contract No. DE-AC02-06CH11357.
This research was partially supported by the Catalyst Design for Decarbonization Center, an Energy Frontier Research Center funded by the U.S. Department of Energy, Office of Science, Basic Energy Sciences under award no. DE-SC0023383 and NSF grant OAC-2514142.

\bibliographystyle{IEEEtran}
\bibliography{refs}

\end{document}

%% file: misc/title_and_authors.tex
\title{%
Steering an Active Learning Workflow Towards Novel Materials Discovery via Queue Prioritization
}

\author{
    \IEEEauthorblockN{
        Marcus Schwarting\IEEEauthorrefmark{1}, Logan Ward\IEEEauthorrefmark{2}, Nathaniel Hudson\IEEEauthorrefmark{1}, Xiaoli Yan\IEEEauthorrefmark{3}, \\
        Ben Blaiszik\IEEEauthorrefmark{2}, Santanu Chaudhuri\IEEEauthorrefmark{3}, 
        Eliu Huerta\IEEEauthorrefmark{2}, Ian Foster\IEEEauthorrefmark{1}
    }
    \IEEEauthorblockA{
        \IEEEauthorrefmark{1}%
        Department of Computer Science, University of Chicago
    }
    \IEEEauthorblockA{
        \IEEEauthorrefmark{2}%
        Data Science \& Learning Division, Argonne National Laboratory
    }
    \IEEEauthorblockA{
        \IEEEauthorrefmark{3}%
        Department of Civil, Materials, and Environmental Engineering, University of Illinois at Chicago
    }
}

%% file: misc/abstract.tex
\begin{abstract}
Generative AI poses both opportunities and risks for solving inverse design problems in the sciences.
Generative tools provide the ability to expand and refine a search space autonomously, but do so at the cost of exploring low-quality regions until sufficiently fine tuned.
Here, we propose a queue prioritization algorithm that combines generative modeling and active learning in the context of a distributed workflow for exploring complex design spaces.
We find that incorporating an active learning model to prioritize top design candidates can prevent a generative AI workflow from expending resources on nonsensical candidates and halt potential generative model decay.
For an existing generative AI workflow for discovering novel molecular structure candidates for carbon capture, our active learning approach significantly increases the number of high-quality candidates identified by the generative model.
We find that, out of 1000 novel candidates, our workflow without active learning can generate an average of 281 high-performing candidates, while our proposed prioritization with active learning can generate an average 604 high-performing candidates.

\end{abstract}

%% file: sections/1_introduction.tex
In recent years, \textit{Generative AI}~(GenAI) models have significantly changed how computational screening workflows are designed and executed.
While initially trained to produce original images and videos, scientists are now using GenAI models to navigate across large, complex, and vastly under-explored search spaces to discover novel materials and designs.
GenAI approaches have augmented computational workflows in domains from mechanical shape optimization~\cite{oh2019deep} to drug discovery~\cite{bian2021generative}. However, introducing GenAI models into these workflows introduce their own challenges.
Generative AI models are trained to emulate the traits of past successful \enquote{candidates} (or outputs), but a significant portion of these novel candidates may be nonsensical.
Without effective candidate prioritization, significant computational resources will be wasted exploring suboptimal regions of design space.
Also, generative models retrained on candidates from suboptimal regions of design space are liable to model decay~\cite{shumailov2023curse}; that is, a steady degradation in the quality of candidates.

In this work, we aim to address this challenge for incorporating generative models in workflows for novel scientific discovery. 
Our proposed queue prioritization algorithm uses an active learning model with a custom acquisition function to prioritize novel candidates produced by a generative model.
Generative models excel at producing novel design candidates, and active learning models excel at selecting promising candidates.
Placed together, these two models play to the strengths of one another to rapidly identify high-performing novel candidates while ignoring unreasonable ones.
The resulting workflow is more efficient at generating novel, diverse, and high-performing candidates.
We further show that, by de-prioritizing unreasonable candidates, active learning can prevent generative models from symptoms of model decay due to iterative retraining.

We demonstrate the efficacy of our proposed solution for a real scientific use case.
For this, we consider a generative workflow that aims to produce porous materials, known as \textit{metal-organic frameworks}~(MOFs), for carbon capture~\cite{yan2025mofa} and apply our queue prioritization algorithm to improve novel material discovery.
MOFs are a versatile class of materials that are relevant to applications in, e.g., gas storage, catalysis, drug delivery, and environmental remediation.
Computational methods for screening chemical space for novel MOFs have predicted half a million stable MOF structures.
Recent high-throughput screening workflows have introduced generative models to suggest novel structures based on previously identified stable MOFs~\cite{yan2025mofa,park2025multi,park2024generative}.
While these models can generate a number of candidates, a large portion are quickly dismissed as chemically infeasible.
Others lack the chemical properties required for specific applications, or fail to achieve the correct balance of competing design priorities.
These challenges make MOF discovery an ideal domain for assessing our proposed combination of generative AI and active learning.

Using this MOF generative workflow, we demonstrate how our queue prioritization algorithm can be cheaply and seamlessly incorporated alongside a generative model to prioritize structure simulation queues, leading to more promising MOF structure candidates.
Specifically, our AL prioritization approach can remove chemically unreasonable structures that might otherwise be used for iterative fine-tuning.
We also demonstrate how multiple candidate properties can be incorporated into our AL prioritization model to match realistic MOF design criteria.
Finally, we show AL can be tuned to incorporate exploration and exploitation objectives as a solution for effectively navigating chemical space while simultaneously training a robust surrogate model for future use.
We find that, out of 1000 novel candidates, our workflow without active learning can generate an average of 281 high-performing candidates, while our proposed prioritization with active learning can generate an average 604 high-performing candidates.
These gains are accomplished with a marginal increase in computational overhead, which we estimate as below 1\% of the overall workflow costs.

%% file: sections/2_related.tex
\subsection{Generative Models}
\label{sec:related_work_gen_ai}

\subsubsection{Applications of Generative Models}
GenAI models have now been integrated into many different domains for multiple applications.
Cheminformatics workflows have been created to generate molecules to satisfy many design criteria, such as drug discovery~\cite{bian2021generative}, proteomics~\cite{john2024bionemo}, polymer design~\cite{kim2023open}, and identifying novel MOFs~\cite{park2025multi}.
The first generative deep learning architectures were \textit{variational autoencoders}~\cite{gomez2018automatic} and \textit{generative adversarial networks}~\cite{prykhodko2019novo}.
These have now given way to normalizing flow and diffusion architectures \cite{xu2022geodiff} and transformer-based models \cite{fabian2020molecular}.
The general approach is to encode molecular structures $m_i \in \Omega$ (where $\Omega$ represents possible compounds) into latent representations $r_i \in \mathbb{R}^n$ to create previously unexplored compounds with desirable properties.
Generative AI models can offer a distinct advantage over brute-force screening when exploring chemical space.

\subsubsection{Disadvantages of Generative Models}
Despite recent advances in generative models, one persisting challenge is ensuring that candidates are sensible.
Generative models, particularly those trained on human-biased datasets, may prioritize regions of design space that are impractical or de-prioritize regions that are attractive \cite{gao2020synthesizability}.
The disconnect between computational predictions and experimental feasibility can lead to wasted resources when screening or validating candidates.

\subsection{Active Learning for Scientific Workflows}
\label{sec:related_work_al}
\subsubsection{Workflow Steering}
Prior to the widespread adoption of generative AI models, many high-throughput screening workflows were steered by AL \cite{graff2021accelerating, reker2015active}. 
AL assumes a small labeled dataset $D_L$ and a large amount of unlabeled data $D_U$, and can access an oracle to provide in-situ labels on samples from $D_U$.
However, simulated experimentation can also be used \cite{chung2019advances}, as is the case here, where the authors use a \textit{Molecular Dynamics}~(MD) simulation code as the oracle.
AL approaches leverage the predictions of a classical supervised method while simultaneously expanding on $D_L$ to refine subsequent model predictions.

\subsubsection{Explore \& Exploit Trade-offs}
A common theme in AL is the trade-off between exploitative and exploratory search \cite{settles2009active}.
Exploitation prioritizes a specific specific scoring function (or functions).
Exploration prioritizes samples from diverse regions of state space, making significant departures from known successful candidates.
An acquisition function that balances exploration and exploitation often outperforms an acquisition function limited to one extreme.
For a given molecular candidate $m$, a supervised model $\mathcal{M}_i$ provides a property prediction $\mathcal{M}_i(m)$ and an uncertainty metric $\sigma_{\mathcal{M}_i}(m)$.
Uncertainty may arise from statistical presuppositions (such as with a Gaussian process regressor) or from ensemble variability (such as with a random forest regressor).
Solely prioritizing the property prediction $\mathcal{M}_i(m)$ is an exploit-only approach, while solely prioritizing the uncertainty $\sigma_{\mathcal{M}_i}(m)$ is an explore-only approach.
If the goal is to maximize a given property, an acquisition function like \textit{upper confidence bound}~(UCB) is a popular choice to prioritize candidates: $\mathit{UCB}(m) = \mathcal{M}_i(m) + \lambda \sigma_{\mathcal{M}_i}(m)$, where $\lambda$ calibrates the emphasis on the exploration objective \cite{li2006confidence, schneider2022silico}.


\subsubsection{Optimizing Queues and Schedulers}
Starting around 2010, AL was first incorporated into novel scheduling algorithms~\cite{yang2011task}.
For applications such as dynamic market transactions~\cite{afeche2013bayesian} and medical triage~\cite{singh2024feature}, the degree of urgency allocated to a given candidate (e.g., a customer or patient) in a priority queue cannot be easily ascertained.
A singly-trained surrogate model that makes predictions on a collection of candidate features can offer an initial benefit for queue prioritization, however incorporating AL retraining or fine-tuning (even at infrequent intervals or solely on recently obtained data) can ensure that a model remains up-to-date.



%% file: sections/3_problem.tex
In this work, we consider workflows that use generative AI to create novel candidates in complex search spaces.
This section outlines a general framework for generative workflows before we discuss how to accelerate navigation through this search space to make novel discoveries.

\subsection{Simulators and Scoring Functions}
We assume a set of scoring functions that, taken together, indicate how viable a candidate~$m$ should be considered for a given application.
We further assume a given scoring function will return a positive number for all candidates, $S_i(m)\in \mathbb{R}_+$, such that a list of candidates can be placed in ascending order with respect to $S_i$.
Without loss of generality, given candidates $m_1$ and $m_2$, suppose that $S_i(m_1) > S_i(m_2)$ implies that $m_1$ is a better candidate than $m_2$ with respect to the scoring function $S_i$.
Some scoring functions may be cheap to compute (for notational convenience, we call these $\{S_a, S_b,\ldots \}$), while others may require significant computational resources to carry out a simulation (call these $\{S_{IS},\ldots \}$).
For our work, we consider that for a computationally expensive scoring function $S_{IS}$ that has been evaluated for a sufficiently large number of candidates, it is possible to train a machine learning surrogate $\mathcal{M}_{IS}$ which approximates the scoring function for a fraction of the compute resources.

\subsection{Training and Results Datasets}
We assume the availability of a labeled pre-training dataset $D_{PT}$ of previously scored candidates, which can serve as training data for a supervised surrogate model as well as for an unsupervised generative model.
This dataset may consist of candidates derived from literature, ab-initio experiments, or computational databases.
We initialize an empty database $D_S$, where we store candidates and their respective scores throughout a workflow run.
Upon completion of the workflow, we denote the subset of high-performing candidates by $D_S^\star = \{m_i \in D_S \; \vert \; S_{IS}(m_i)>t_{IS};\; S_a(m_i) > t_a, ... \}$, where $t_{(\cdot)} \in \mathbb{R}_+$ represents a fixed threshold for each scoring function $S_i$ that a candidate must exceed.

\subsection{Generative Model}
We assume we are given a pre-trained generator $\mathcal{G}:\mathbb{R}^n \to \Omega$ (trained on $D_{PT}$, or some super-set thereof) that returns novel candidates $\{ m_1, \ldots\} \subset \Omega$.
The definition of a generator function in our work is general enough that it can be implemented by most GenAI models, such as denoising diffusion models~\cite{ho2020denoising}.
Candidates returned by a generator are added to a queue $Q_{GL}$ in the order they were generated.
In the absence of prioritization on $Q_{GL}$, we assume that candidates will be scored with all scoring functions in the order they were generated.
We further assume that $\mathcal{G}$ can be iteratively \textit{fine-tuned} on a subset of previously generated high-performing candidates $D_S^\star \subset D_S$, with the intention of driving the generator to produce a greater fraction of high-scoring candidates over time.


\subsection{Workflow Objectives}
Given the current workflow setup, our primary objective is to maximize the fraction of novel candidates that satisfy a combined scoring threshold.
Suppose $D_{S,T}^* \subset D_{S,T}$ is the set of candidates that meet all scoring thresholds after $T$ candidates ($T = \vert D_{S,T}\vert$) have been generated, post-processed, and simulated.
Further, we assume that all candidates are unique; that is, $m_i \neq m_j\;(\forall m_i,m_j \in D_{S,T}; i \neq j)$.
We also aim to have all candidates in $D_{S,T}$ to be novel, so $D_{S,T} \cap D_{PT} = \emptyset\; (\forall t = 1, \ldots, T)$. 
Let $R_T = \vert D_{S,T}^* \vert / \vert D_T\vert \in [0,1]$ be the fraction of satisfactory novel candidates.
While it cannot be guaranteed that $R_T$ increases monotonically as $T \to \infty$, since $\mathcal{G}$ will still generate unsatisfactory candidates on occasion, our objective would be to improve $\mathcal{G}$ to the point where $R_T \to 1$ as $T\to \infty$.
Note that, in order to avoid an outcome where $\mathcal{G}$ generates a variety of candidates that differ in trivial ways, it is important to include a scoring function in $D_S^*$ that promotes diversity amongst high-performing candidates.
As a secondary objective, we also wish to construct an effective surrogate model $\mathcal{M}_{IS}$ for predicting expensive simulated properties, as this further improves the queue prioritization algorithm and ultimately helps increase $R_T$.
We might write this as minimizing a \textit{root-mean squared error}~(RMSE) between predicted and actual properties, expressed as seen below in \autoref{eq:rmse}:

\begin{equation}
\footnotesize
    \text{RMSE} =
    \sqrt
    { 
        \frac{1}{\vert D_{PT} \cup D_S \vert } \sum_{m_i \in D_{PT} \cup D_S} (S_{IS}(m_i) - \mathcal{M}_{IS}(m_i))^2
    }.
\label{eq:rmse}
\end{equation}



%% file: sections/4_solution.tex
\begin{figure}
    \centering
    \includegraphics[width=0.98\columnwidth]{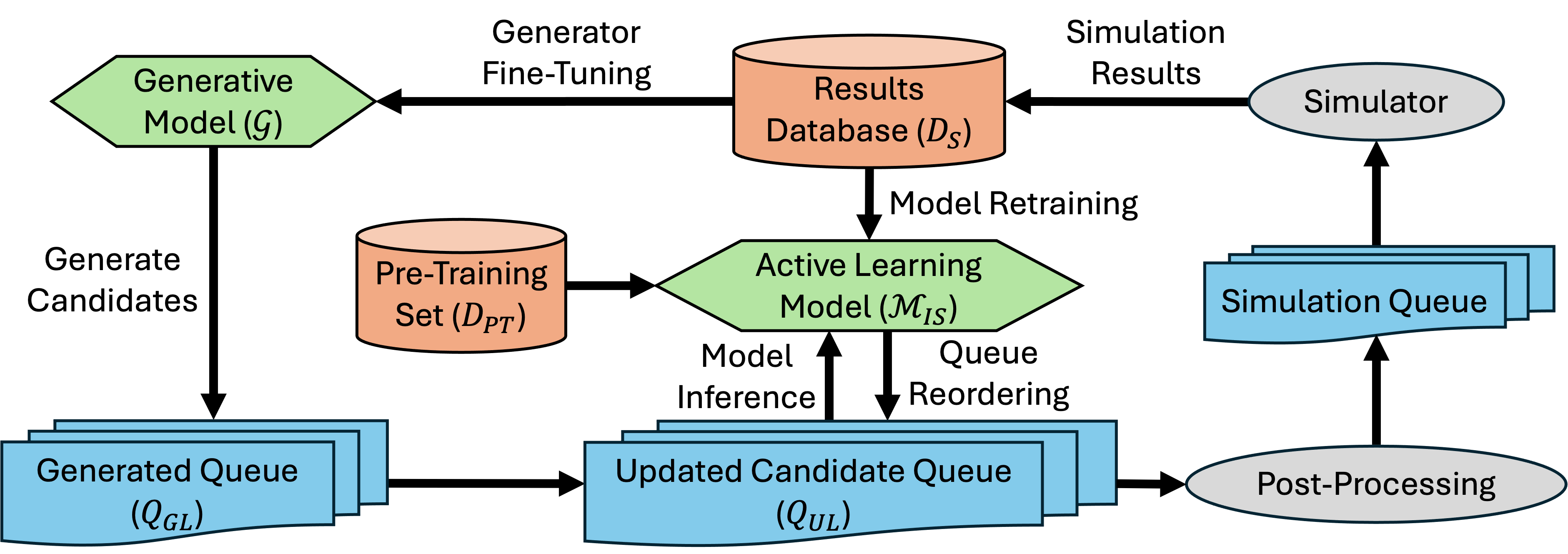}
    \caption{Flow chart of a generative model workflow with AL for queue prioritization.}
    \label{fig:mofa_generic_workflow}
\end{figure}

Given a workflow as described in \autoref{sec:problem_setup}, unrealistic candidates are frequently scored only to be identified later as unusable.
We propose a queue prioritization algorithm which uses AL prioritize candidates that are most likely to be high-scoring, which synergizes with the generative model to promote the creation of more high-scoring candidates.
We train an initial surrogate model, $\mathcal{M}_{IS}$, that starts with a dataset of candidates with pre-computed features, $D_{PT}$, and model retraining occurs as new candidates are simulated.
Our supervised task is therefore to learn a function $\mathcal{M}_{IS}:\mathbb{R}^{d} \to \mathbb{R}_+$.
As novel candidates $\{m_1,...\}$ are created by the generative model~$\mathcal{G}$, they are stored in a non-prioritized queue $Q_{GL}$.
For a given candidate, we combine scoring predictions from $\mathcal{M}_{IS}(m_i)$, in conjunction with scoring functions $\{S_a(m_i), S_b(m_i), ... \}$, into an acquisition function $\mathcal{A}(m_i) \in \mathbb{R}_+$.
We then reorder candidates in a new prioritized queue, $Q_{UL}$, such that candidates ranking the highest according to the acquisition function are the first to be evaluated.
\autoref{fig:mofa_generic_workflow} shows how our queue prioritization algorithm can be incorporated into a generative model screening workflow to reorder candidates from the $Q_{GL}$ queue to improve the quality of candidates that are scored by the simulator.


For our surrogate model, $\mathcal{M}_{IS}$, we use an XGBoost regressor in our workflow for several reasons: 
\textit{(i)}~training and inference are both quick and trivially parallelizable, 
\textit{(ii)}~learned weights can be represented concisely and quickly saved to, and loaded from, disk when necessary, 
and 
\textit{(iii)}~a calibrated ensemble uncertainty metric $\sigma_{IS}$ can be easily ascertained by running and aggregating predictions across individual trees.
But, it should be noted that the framework for our approach is general enough to support other regressors or classifiers for the surrogate model.

\input{algorithms/active_learning}

\paragraph{Algorithm Review}
Pseudocode for our proposed prioritization method is presented in \autoref{algo:al_queue_reprioritization}.
We begin with a dataset of pre-training candidates $D_{PT}$, a model $\mathcal{M}_{IS}$ trained on $D_{PT}$ to predict $S_{IS}$, and an acquisition function $\mathcal{A}$.
We initialize a queue $Q_{GL}$ to hold candidates created by $\mathcal{G}$, a queue $Q_{UL}$ to hold candidates ranked according to $\mathcal{A}$, and an empty database $D_S$ to store simulation results (line 1).
Over the duration of the workflow, when simulations complete and are stored in $D_S$, our model $M_{IS}$ is retrained on both $D_{PT}$ and $D_S$.
After our model has been updated, we run inference on all candidates $m_b$ on queues $Q_{GL}$ and $Q_{UL}$, which yields score predictions $\mathcal{M}_{IS}(m_b)$ and corresponding score uncertainties $\sigma_{\mathcal{M}_{IS}}(m_b)$.
These predictions and uncertainties are incorporated into the acquisition function $\mathcal{A}$, which is used to prioritize the $Q_{UL}$ queue with all new candidates.
These are post-processed starting from the top-scoring candidates according to $\mathcal{A}$.
As simulations are completed, the results are stored in $D_S$.
The workflow halts once a fixed number of novel candidates have been simulated (in our case, 1000 candidates).


%% file: algorithms/active_learning.tex
\begin{algorithm}[t]
\caption{Proposed AL Queue Prioritization}
\label{algo:al_queue_reprioritization}
\small

\KwIn{Dataset of pre-computed candidates $D_{PT} = \{(m_a, S_{IS}(m_a)), ... \}$, initial prediction model $\mathcal{M}_{IS}$ (pre-trained on $D_{PT}$), acquisition function $\mathcal{A}$, $N$ total candidates to generate.}
\KwOut{Final trained model $\mathcal{M}_{IS}$ and top candidates $D_S^\star$.}

Initialize Initial queue $Q_{GL} = []$, Updated queue $Q_{UL} = []$, dataset of simulated candidates $D_S = \{ \}$\;


\While {$\vert D_S \vert < N$}
{
    \If{new data on $D_S$ since training $M_{IS}$}{
        Train $\mathcal{M}_{IS}^\dagger$ on $D_{PT} \cup D_S$\; Replace stale model $\mathcal{M}_{IS} \leftarrow \mathcal{M}_{IS}^{\dagger}$\;
    }
    
    \tcp{Calculate metrics for prioritization}
    \ForEach{candidate $m_j \in Q_{GL} \cup Q_{UL}$}{
        Compute model predictions and uncertainties $\mathcal{M}_{IS}(m_j),\sigma_{\mathcal{M}_{IS}}(m_j)$\;
        
        Compute acquisitions $\mathcal{A}_{m_j}$ with $\mathcal{M}_{IS}(m_j),\sigma_{\mathcal{M}_{IS}}(m_j)$\;
    }

    Prioritize queue via $\textsc{Sort}(Q_{UL})$ in ascending order according to $\mathcal{A}(\cdot)$\;

    \lWhile{simulation is not finished}{wait}
    Add results, via $D_S \leftarrow D_S \cup \{(m_c, S_{IS}(m_c)), \ldots\}$\;

    \lWhile{Post-processing is not yet completed}{wait}
    Draw top candidate, $m_d \gets \mathit{pop}(Q_{UL} \vert \mathcal{A})$\;
}
\end{algorithm}

%% file: sections/5_mof_application.tex
Here, we discuss the application we use to demonstrate the efficacy of our proposed prioritization for novel discovery with active learning. 
We apply our proposed queue prioritization algorithm for AL to the task of discovering novel \textit{Metal-Organic Frameworks}~(MOFs).
We use a pre-existing workflow for generating novel MOFs which matches the problem setup~\cite{yan2025mofa}, and describe how our proposed solution can be adapted to this specific use case.

\subsection{MOF Construction}
\label{sec:mof_construction}
We briefly describe the chemical details of MOFs that are relevant to this work.
MOFs are composed of two parts: clusters of metal atoms laid out on a three-dimensional grid, and \textit{linker} molecules spanning the grid to connect the metal atoms.
While MOFs can come in a variety of shapes, we limit our current search to cubic MOFs.
Although the chemical space of possible metal atoms is small, the chemical space of linkers is extremely large.
For our workflow, we keep the metal atom fixed (as zinc) and search the space of organic linkers.
We assume that to form a viable simulated MOF, a suitable linker must satisfy two primary constraints.
MOFs must contain \textit{coordination sites} at either end to bind to the metal atoms.
We accommodate this constraint by selecting two common coordination sites which are fixed at either end of all linkers.
Even with this additional constraint within chemical space, linkers can still vary widely.

\begin{figure*}[t]
    \centering
    \subfloat[][Example MOF from the HMOF dataset~\cite{wilmer2012structure}.]{
        \includegraphics[width=0.33\linewidth]{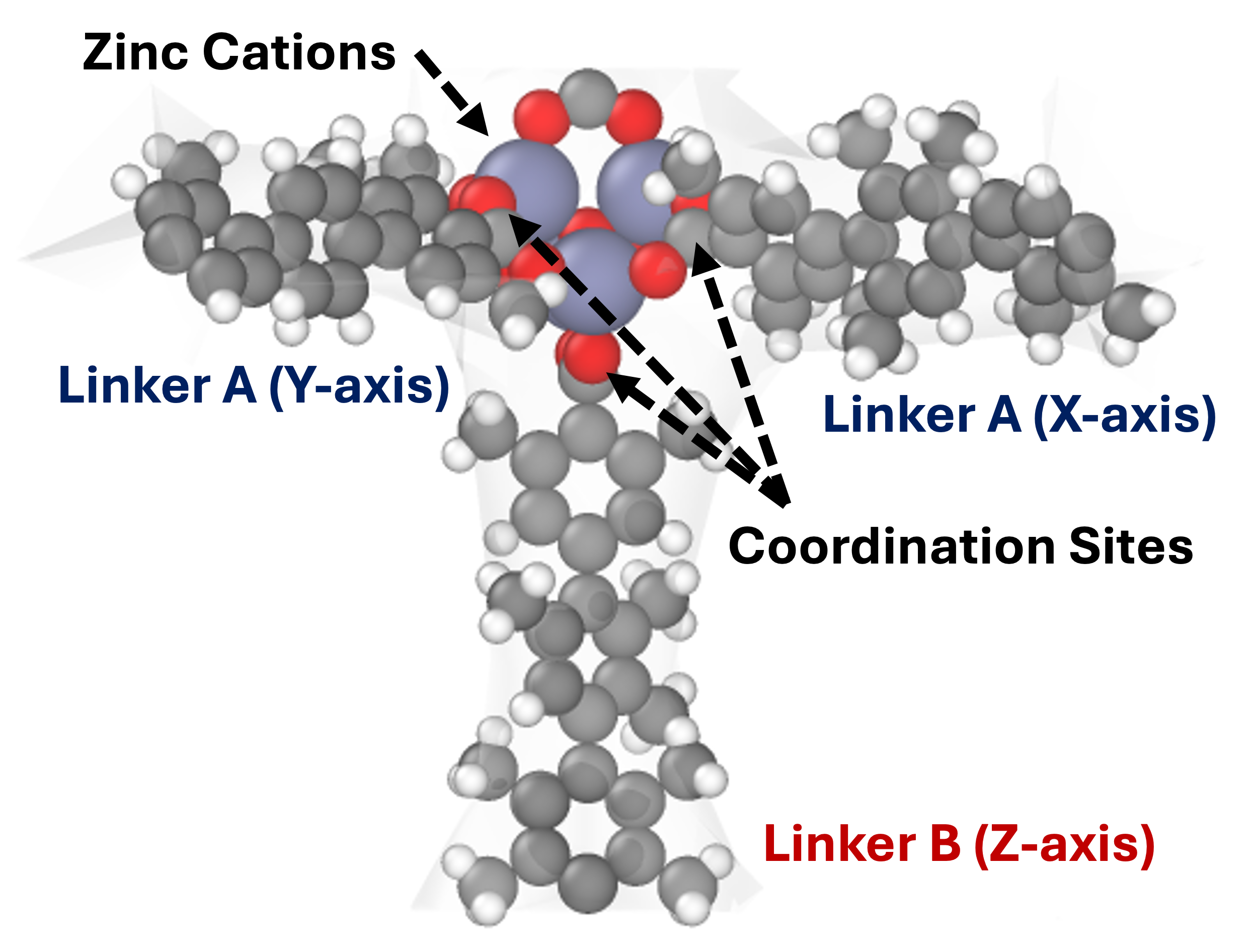}
        \label{fig:mofa_example_workflow_mof}
    }
    \hfill
    \subfloat[][Diagram the of MOF screening workflow]{
        \includegraphics[width=0.63\linewidth]{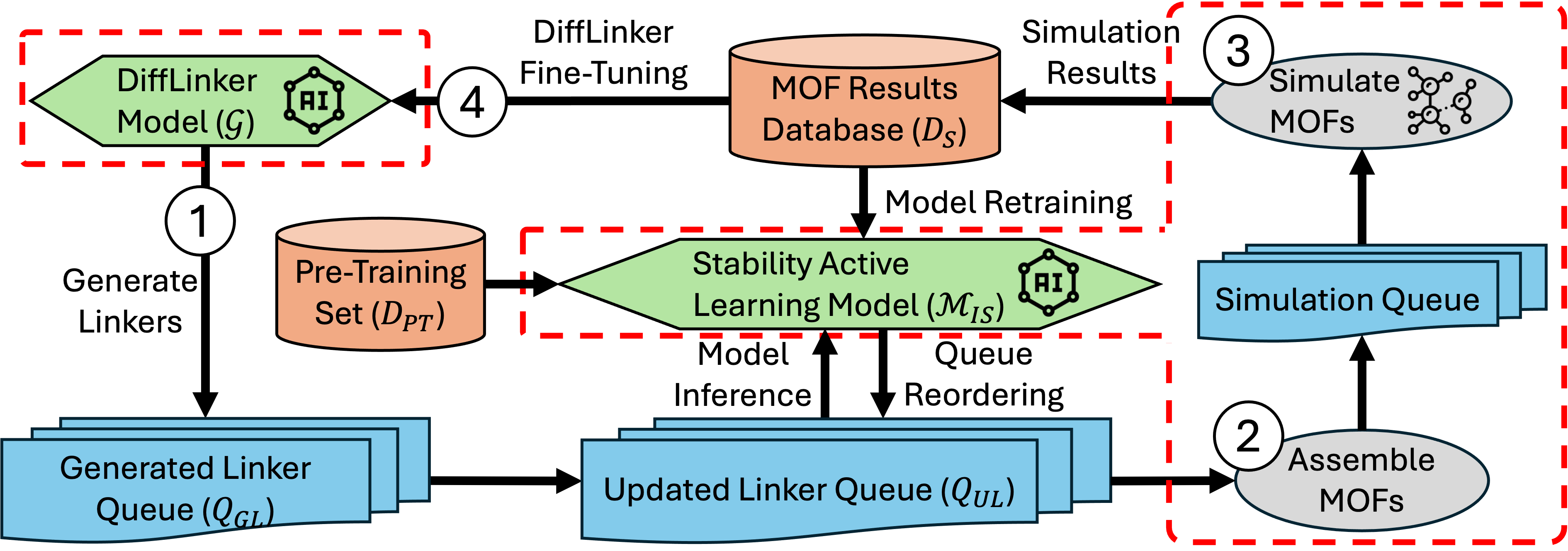}
        \label{fig:mofa_example_workflow_diagram}
    }
    \caption{Example of MOF structure and diagram of MOF screening workflow with the AL model iteratively reordering the updated linker queue.
    Components outlined in red-dashed boxes incorporate MOF-specific implementation details.
    }
    \label{fig:mofa_example_workflow}
\end{figure*}

When linkers are assembled into a MOF, the overall attractions and repulsions between linkers and metal atoms must be balanced to maintain the overall structure.
We measure the stability of a linker~$m_i$ in terms of the unitless scoring function of \textit{Internal Strain}, $S_{IS}(m_i)$.
Although internal strain is a property of an assembled MOF, we assume that the linkers equally contribute to stability, and thus approximate $S_{IS}(m_i)$ for a given linker $m_i$.
We define a \textit{stable MOF} as a MOF with linkers $(m_i,...)$ such that $\forall m_i,\; S_{IS}(m_i)<0.25$, which is considered a reasonable threshold of internal strain \cite{yan2025mofa}.
Only a handful of our generated MOFs clear this threshold.
Below an internal strain threshold of $<0.25$, MOFs can be considered equivalently stable to one another.
\autoref{fig:mofa_example_workflow_mof} shows one such stable MOF with components labeled accordingly.

In addition to stability $S_{IS}$, the viability of a MOF structure can be considered from two alternative perspectives, which are far less computationally expensive to determine.
First, we can determine the \textit{Synthesizability Assessment Score}~(SAScore) of the two linkers comprising the MOF \cite{ertl2009estimation}.
For a given organic molecule $m$, the SAScore~$S_{SA}(m)\in[0,1]$ is a normalized float metric that estimates the likelihood that it can be synthesized (scoring 0) or cannot be synthesized (scoring 1).
Second, we can determine how similar a given linker is to previously identified linkers from the hypothetical-MOF (HMOF) dataset \cite{wilmer2012structure} using a normalized Tanimoto float metric \cite{bajusz2015tanimoto}.
For two molecules $m_1,m_2$ with functional groups $\mu_1 = \{ \mu_{1,1},..., \mu_{q,1}\}$ and $\mu_2 = \{ \mu_{1,2}, ..., \mu_{r,2}\}$, the Tanimoto distance is defined as
$$d_T(m_1,m_2) = 1-\frac{\vert \mu_1 \cap \mu_2 \vert}{\vert \mu_1 \vert + \vert \mu_2 \vert - \vert \mu_1 \cap \mu_2 \vert}.$$

A Tanimoto distance $d_T(m_1,m_2)$ of 0 implies two molecules $m_1,m_2$ are identical, whereas a score of 1 implies the two molecules' dissimilarity is maximum.
For each novel linker proposed, we take the Tanimoto score of the most similar linker in the HMOF dataset, or $S_T(m)=\min_{m_i \in \text{HMOF}}\{ d_T(m,m_i) \}$.
In the context of this workflow, the ideal MOF would be highly stable ($S_{IS}\simeq 0$) and be composed of linkers $m_1,m_2$ that are experimentally synthesizable ($S_{SA}(m_1),S_{SA}(m_2) \simeq 0$) and resemble existing HMOF linkers ($S_T(m_1),S_T(m_2)\simeq 0$).


\subsection{Simulation Workflow}
For our simulation workflow, we are interested in exploring the chemical space of possible candidate linkers.
We represent linkers using an \textit{RDKit embedding} of 38 features \cite{landrum2013rdkit}.
Our supervised task is therefore to learn a function $f:\mathbb{R}^{38} \to S_{IS} \in \mathbb{R}_{+}$.
Our workflow for identifying novel MOFs can be broken into four tasks, as shown in \autoref{fig:mofa_example_workflow_diagram}.
Complete implementation details of the MOF workflow, as well as metrics for tests performed on HPC systems, see \cite{yan2025mofa}.

\paragraph{Generate Linkers}
We generate linker candidates by using a fine-tuned DiffLinker model \cite{igashov2024equivariant}
which
is a diffusion model for molecule design---originally trained for drug discovery.
For our workflow, we start with a fine-tuned DiffLinker model trained on a subset of the HMOF dataset~\cite{wilmer2012structure}.
\paragraph{Assemble MOFs}
MOFs are assembled by placing metal atoms at the ``origin'' of a three-dimensional box, then surrounding these atoms with linkers along orthogonal axes in an orientation so the coordination site is near the metal atoms.
\paragraph{Simulate MOFs}
The internal strain of a MOF is an important property when it comes to determining whether a MOF is likely to be experimentally synthesizable: too high a value, and the MOF, if it could be assembled at all, is likely to collapse in on itself.
Thus we employ the LAMMPS MD package \cite{thompson2022lammps} to simulate this property, and reject MOFs for which its value is too high.
This step is the most computationally expensive.
\paragraph{DiffLinker Fine-Tuning}
A small fraction of MD-simulated MOFs prove sufficiently stable to be deemed experimentally feasible.
We seek to increase the faction of stable MOFs by iteratively refining the DiffLinker model on these successful candidates.
For every new MOF candidate below a set strain threshold, we fine-tune the DiffLinker model on the subset of all previously identified successful linkers \cite{park2024generative}.

%% file: sections/6_results.tex
Here, we present and discuss the results of our queue prioritization in the context of generating novel MOFs.
In executing our workflow, we set the number of candidates to generate in \autoref{algo:al_queue_reprioritization} to $N=1000$.
For CPU-heavy operations, we use 48 Intel Xeon CPUs to run MD simulations.
We use one Intel Xeon CPU to perform MOF structure assembly from ligands, and one Intel Xeon CPU to carry out AL model retraining/inference.
For GPU-heavy operations, we use one Nvidia RTX-2080 GPU for DiffLinker fine-tuning/inference.

\subsection{Baseline Data Analysis}
We first compare simulated linkers generated by the DiffLinker model to those from a known theoretical database.
We use the linkers from the HMOF dataset of around 138K MOF structures, of which we select a random subset of 3.3K structures to calculate stability $S_{IS}$.
We also aggregate results of previous runs of our workflow without AL, which yielded over 49K unique MOFs with widely varying stabilities.
These runs were performed on the Argonne Polaris supercomputer, so we refer to this collection of structures as the \textit{Polaris dataset}.
We see that the DiffLinker model generates a variety of linkers in the Polaris dataset that are structurally diverse compared to the HMOF linkers.
However, the wide-ranging structural diversity produced by DiffLinker leads to a large number of low-stability linkers taking significant compute resources.

The linkers in the Polaris dataset represent a wide sample of chemical space.
We see that HMOF linkers are far more homogeneous (average pairwise Tanimoto distance of 0.15), while Polaris linkers are more structurally diverse (average pairwise Tanimoto distance of 0.36).
Likewise, Polaris and HMOF linkers tend to be quite dissimilar (average $S_T$ of 0.49).


\subsection{Baseline Data Reordering}
\label{sec:results_baseline_data_reorder}
We start by testing the effectiveness of our queue prioritization algorithm with a simulated search using known linkers and stabilities.
We use the subset of HMOF linkers with known stabilities ($\sim$3.3K total unique linkers), along with the Polaris linkers dataset ($\sim$36K total unique linkers).
From an initial set of 200 random linkers, we draw batches of 200 linkers until all samples have been \enquote{acquired.}
At each iteration we retrain our surrogate model $\mathcal{M}_{IS}$ on all acquired linkers.

We express the exploitation/exploration trade-off by using an upper confidence bound~(UCB) acquisition, $UCB_\lambda(m) = \mathcal{M}_{IS}(m) - \lambda \sigma_{\mathcal{M}_{IS}}(m)$.
Increasing the value of $\lambda$ in the acquisition function places greater priority on an exploration objective.
We validate exploitation performance by checking the cumulative number of stable MOFs, and we validate exploration performance by measuring the model predicted internal strain RMSE on a hold-out set of $\sim$3K linkers.
We average these performance metrics across three distinct reordering trials (starting with a different initial random selection each for each trial).
These results are compared against random re-orderings of MOF structures, with model retraining occurring at intervals of 200 linkers.
We list these training set reordering tasks in the first row of \autoref{tab:all_workflow_tests}.

We find that that our AL model can reorder the queue of MOFs to better prioritize overarching exploration and exploitation objectives, as compared to a random selection as a baseline.
\autoref{fig:mofa_al_out_of_loop} shows how AL using an exploit-only objective can outperform other approaches at rapidly selecting the most stable MOFs, with the trade-off that the AL model will yield worse $S_{IS}$ predictions overall.
Conversely, an explore-only objective can still outperform a purely random baseline for selecting stable MOFs, but excels at training a model that yields better $S_{IS}$ predictions with fewer training simulations.
Between these two extreme strategies is UCB for weak exploration ($\lambda=0.1$) and strong exploration ($\lambda=2.0$).

The performance of weak and strong exploration is nearly identical, and leads to excellent outcomes for model training to reduce $S_{IS}$ prediction RMSE.
This is because a pure exploration acquisition function has an incentive to sample outliers in the embedding space, which are less likely to be stable MOFs and are less helpful for predicting $S_{IS}$ on samples in the holdout set.
Weak and strong exploration with UCB offer a suitable middle ground: the supervised model can make better $S_{IS}$ predictions with far fewer samples, while pointing to more stable structures throughout the reordering process.
The overall effect of incorporating an objective with an exploration/exploitation trade-off is to moderate a generative model by reducing the computational expense of simulating materials that are, in all likelihood, nonsensical.

\input{figs/tex/fixed_ordering}

\input{tables/test_details}

\input{figs/tex/mofa_retraining_fraction}

\subsection{Preventing Model Decay}
Retraining the DiffLinker model is important for improving the quality of generated linkers, however what percentage of top linkers that ought to be used for fine-tuning is not obvious.
If too few top linkers are used for fine-tuning, the model will be gradually adjusted to produce better MOFs, and in the meantime will waste significant computational resources generating, assembling, and simulating unstable structures.
If too many linkers are used for fine-tuning, the model could enter a spiral of decay where training on generations of unstable structures begets future generations of even more unstable structures \cite{shumailov2023curse}.
To understand these extremes, we run our generation workflow to produce a total of 1000 novel MOFs, and fine-tune the DiffLinker model on a percentage of top generated ligands: the top 10\%, 50\%, or 90\% of linkers.
By fine-tuning with these different percentages of samples, we aim to identify a middle ground between these two fine-tuning extrema.
We perform these fine-tuning tests both with and without our proposed AL queue reordering with an exploit-only objective, as shown in \autoref{tab:all_workflow_tests}.
We supply the initial AL model with the pre-training dataset $D_{PT}$ consisting of all HMOF and Polaris linkers with available stability calculations (that is, linkers used in Section \ref{sec:results_baseline_data_reorder}).

We find that, regardless of the fine-tuning fraction, incorporating AL for queue reordering is an effective, computationally efficient way to improve the quality of linkers generated by the fine-tuned DiffLinker model.
\autoref{fig:mofa_al_retrain} shows how, with 90\% of top candidates used for DiffLinker fine-tuning, model decay can lead to higher average internal strains and fewer stable MOFs generated.
However, negative effects of model decay can be offset by incorporating AL for queue reordering.
In this way, the AL model acts as an regulatory intermediate for the generative AI model: de-prioritizing nonsensical linkers while prioritizing those most likely to succeed.
The result of this intermediate prioritization is the successful identification of around 30\% more novel MOFs at the expense of a marginal fixed increase in compute cost.

\subsection{Multi-Objective Prioritization}
Rather than focusing solely on stability ($S_{IS}$) for queue prioritization, we consider modifying with multiple objectives discussed in Section \ref{sec:mof_construction}; namely, HMOF-Tanimoto similarity ($S_T$) and synthesizability score ($S_{SA}$).
Using these additional objectives, we identify linkers that are both more stable and more synthesizable.
We construct five acquisition functions via a linear combination of $S_{IS}$, $S_T$, and $S_{SA}$, and run our generation workflow to produce a total of 1000 novel MOFs, supplying our initial AL model with the $D_{PT}$ pre-training dataset.
We validate workflow performance by capturing a windowed average of novel linker $S_{T}$ and $S_{SA}$ and by counting the cumulative number of stable MOF candidates.
The details of each of these tests are listed in \autoref{tab:all_workflow_tests}.

Using acquisition functions that combine stability ($S_{IS}$) with synthesizability ($S_{SA}$) and HMOF Tanimoto similarity ($S_{T}$), we find that queue re-prioritization can effective in a multiple objective scenario.
\autoref{fig:queue_reordering_st} and \autoref{fig:queue_reordering_ssa} shows the effect of acquisition functions transitioning from no emphasis on $S_{SA}$ and $S_T$ to using these features as a sole focus.
Re-prioritizing using $S_{SA}$, we find that most DiffLinker candidates are already likely to be experimentally synthesized, with $S_{SA}$ averaging below 0.1 in the range normalized [0,1] (where $S_{SA}=0$ suggests the linker is synthesizable).
Therefore, slight improvements in $S_{SA}$ have little downstream impact on a multi-objective acquisition function.
When $S_{SA}$ is solely prioritized, we see a slight decrease in the $S_{SA}$ scores but a significant drop in the total number of novel stable MOFs.
When prioritizing using $S_T$, we find that identifying novel linkers similar to HMOF leads to more stable MOFs, but does not guarantee that these candidates are experimentally viable.

\input{figs/tex/queue_reordering}


\subsection{Explore \& Exploit Tuning}
In the worst case, using an exploit-only acquisition function can lead to our workflow prioritizing linkers that are almost the same as known stable linkers, which is of far less interest than highly novel stable linkers.
We propose combating this sameness incorporating an exploration objective, as measured by the uncertainty quantification $\sigma_{\mathcal{M}_{IS}}$ of the $\mathcal{M}_{IS}$ model.
We use an \textit{upper confidence bound}~(UCB) acquisition function, as described in Section \ref{sec:results_baseline_data_reorder}, with $\lambda = 0.1$ for ``weak'' exploration and $\lambda = 2.0$ for ``stron'' exploration.
We run our generation workflow to produce a total of 1000 novel MOFs, supplying our initial AL model with the $D_{PT}$ pre-training dataset.
The details of each of these tests are listed in \autoref{tab:all_workflow_tests}.

We find that we can prioritize the linker queue using AL with an acquisition function that balances exploration vs.\ exploitation objectives, leading to more promising MOF structures than acquisition functions that focus on one extreme.
With a UCB acquisition function, we find that setting $\lambda=0.1$ for a weak exploration priority has dual benefits that surpass a pure exploitation objective.
\autoref{fig:queue_reordering_st} shows that this AL model achieves the lowest $S_{IS}$ prediction RMSE while simultaneously returning the greatest number of stable MOFs.
Conversely, we find that a pure exploration acquisition function leads to dual drawbacks.
This AL model achieves far fewer stable MOFs, which is expected since no weight is given to prioritizing low internal strain.
However, this AL model also fails to be a useful predictor of $S_{IS}$.
This is due to the exploration objective prioritizing DiffLinker-generated outlier candidates in far-flung regions of chemical space where few, if any, existing nearby data points have been captured.
Almost none of these outlier candidates are worth the computational effort to assemble and simulate, and the results of these simulations are rarely chemically meaningful.
While it is not favorable to entirely exclude under-explored regions of chemical space, it should be understood that exploring these regions represents a high-risk/high-reward proposition.
Because chemical space is so vast, there are a wide range of viable MOFs within a relatively narrow subspace.
This narrow subspace is best navigated using a GenAI model with a surrogate AL model to prioritize screening measurements.

\input{figs/tex/al_explore_exploit_tradeoff}

\subsection{Active Learning Compute Costs}
When our workflow is run on 50 Intel Xeon CPUs, $\sim$30\% of a single CPU is required to carry out all AL retraining and inference operations.
To generate 1000 novel MOFs, all CPUs are engaged for roughly two hours.
During this period, AL queue re-prioritization is triggered between 120 and 130 times (depending on the asynchronous simulation completion).
For each trigger, the entire AL queue prioritization process on a single CPU takes an average of 16 seconds.
Overall, AL queue re-prioritization engages a single CPU for about 30\% of the overall runtime.
In other words, the addition of AL constitutes an overall increase of $\sim$0.6\% in compute resources.
In Table \ref{tab:latency_steps}, we present a breakdown of the compute resources used by each workflow step.
For this marginal increase in computational overhead, our AL method provides a significant improvement in the number of stable MOFs identified.
Without significant loss of generality, we expect AL prioritization to remain a marginal expense as the resource allocation for MD simulations increases.
Our tests are defined with a resource allocation that is far from a tipping point where $M_{IS}$ retraining for every completed simulation becomes a latency bottleneck.
In practice, if such a point is reached, our solution is to simply retrain $M_{IS}$ at batch intervals of completed simulations.

\input{tables/utilization}

%% file: figs/tex/fixed_ordering.tex
    

\begin{figure}
    \centering
    \includegraphics[width=\linewidth]{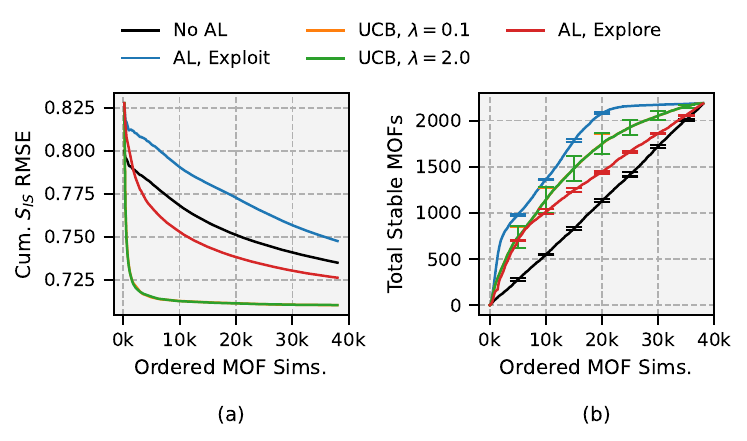}
    \caption{Reordering tests with AL driven by exploration/exploitation, and mixed explore/exploit objectives (compared to a random baseline).
    (a) RMSE of a linker hold-out set with different queue ordering priorities.
    (b) Total \# of identified stable MOFs with different queue ordering priorities.
    }
    \label{fig:mofa_al_out_of_loop}
\end{figure}

%% file: tables/test_details.tex

\begin{table}[t]
\small
\centering
\caption{All AL and control test details, both with queue reordering and application workflow integration.}
\label{tab:all_workflow_tests}
\resizebox{0.48\textwidth}{!}{
\begin{tabular}{
    p{0.2\linewidth}
    p{0.3\linewidth}
    p{0.1\linewidth}
    p{0.1\linewidth}
    p{0.3\linewidth}
}
\toprule
    \textbf{Test Type} & 
    \textbf{Test Name} & 
    \textbf{FT Frac.}  & \textbf{AL?} & 
    \textbf{Prioritization ($\mathcal{A}$)} \\ 
\midrule
    \multirow{5}{*}{\shortstack[c]{Training Set\\Only}} 
        & Random Selection & - & No & $\sim U(0,1)$ \\
        & Exploit Only & - & Yes & $S_{IS}$ \\
         & UCB, Small Explore & - & Yes & $S_{IS} + \frac{1}{10}\sigma_{IS}$ \\
         & UCB, Large Explore & - & Yes & $S_{IS} + 2\sigma_{IS}$ \\
         & Explore Only & - & Yes & $\sigma_{IS}$ \\

\midrule
\midrule

    \multirow{2}{*}{\shortstack[c]{Primary Tests}} 
        & Basic Control & 0.5 & No & None \\
        & Basic AL & 0.5 & Yes & $S_{IS}$ \\

\midrule
    \multirow{4}{*}{\shortstack[c]{Fine-Tune\\Fraction Tests}} 
        & Control Small Frac. & 0.1 & No & None \\
        & Control Large Frac. & 0.9 & No & None \\
        & AL Small Frac. & 0.1 & Yes & $S_{IS}$ \\ 
        & AL Large Frac. & 0.9 & Yes & $S_{IS}$ \\ 

\midrule
    \multirow{5}{*}{\shortstack[c]{Alternative Acq.\\Functions}} 
        & Only $S_{SA}$ & 0.5 & No & $S_{SA}$ \\
        & Only $S_T$ & 0.5 & No & $S_{T}$ \\
        & $S_{IS}$ and $S_{SA}$ & 0.5 & Yes & $\frac{1}{2} S_{IS}+ \frac{1}{2} S_{SA}$ \\
        & $S_{IS}$ and $S_T$ & 0.5 & Yes & $\frac{1}{2} S_{IS} + \frac{1}{2} S_{T}$ \\
        & $S_{IS}$, $S_{SA}$, and $S_T$ & 0.5 & Yes & $\frac{1}{3} S_{IS} + \frac{1}{3} S_T + \frac{1}{3} S_{SA}$ \\ 

\midrule     
    \multirow{3}{*}{\shortstack[c]{Explore/Exploit\\Tradeoffs}} 
        & UCB, Small Explore & 0.5 & Yes & $S_{IS} + \frac{1}{10}\sigma_{IS}$ \\
        & UCB, Large Explore & 0.5 & Yes & $S_{IS} + 2\sigma_{IS}$ \\
        & Explore Only & 0.5 & Yes & $\sigma_{IS}$ \\
\bottomrule
\end{tabular}
}
\end{table}

%% file: figs/tex/mofa_retraining_fraction.tex

\begin{figure*}
    \includegraphics[width=\linewidth]{
        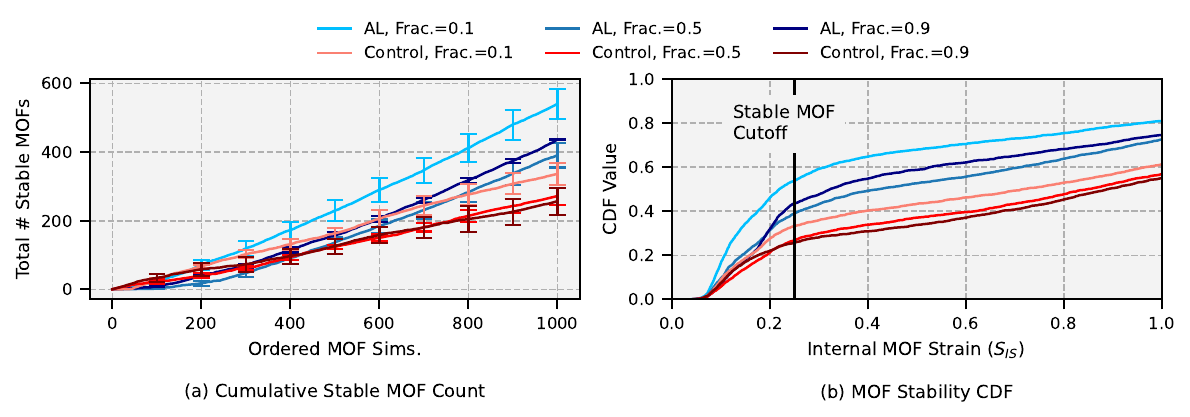
    }
    \caption{Workflow runs with and without AL, shown with varying DiffLinker fine-tuning on a percentage (10\%, 50\%, and 90\%) of the most stable linkers identified.
    (a) Total count of novel stable MOFs during various workflow runs.
    (b) Cumulative proportion of MOFs with corresponding stabilities across various workflow runs.}
    \label{fig:mofa_al_retrain}
\end{figure*}

%% file: figs/tex/queue_reordering.tex
\begin{figure}
   \centering
   \includegraphics[width=\linewidth]{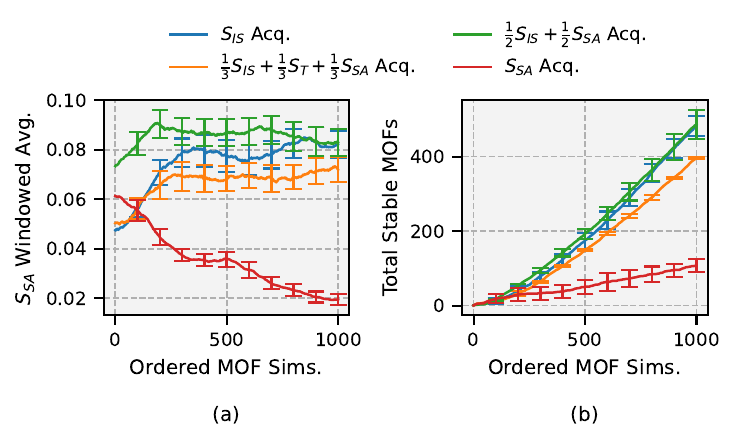}
   \caption{Queue reordering AL tests for alternative acquisition functions with emphasis on SAScore~$S_{SA}$. (a) The averaged $S_{SA}$ over the simulations. (b) The total number of discovered stable MOFs over the simulations.}
   \label{fig:queue_reordering_ssa}
\end{figure}

\begin{figure}
   \centering
   \includegraphics[width=\linewidth]{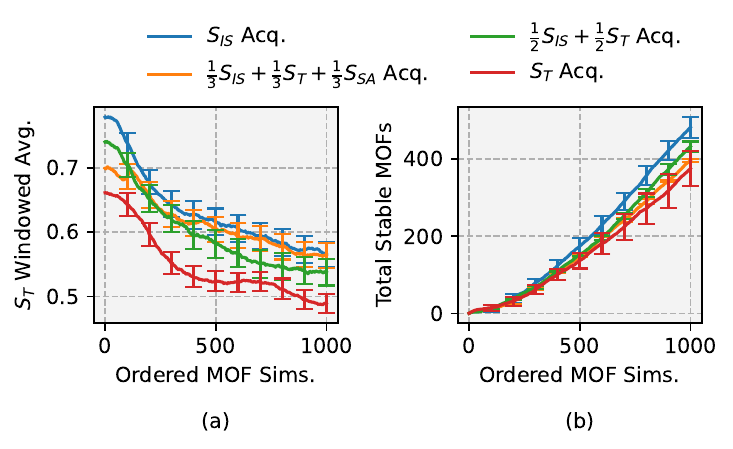}
   \caption{Queue reordering AL tests for alternative acquisition functions with emphasis on HMOF-Tanimoto Similarity~$S_{T}$. (a) The averaged $S_{T}$ over the simulations. (b) The total number of discovered stable MOFs over the simulations.}
   \label{fig:queue_reordering_st}
\end{figure}

%% file: figs/tex/al_explore_exploit_tradeoff.tex

\begin{figure}
    \includegraphics[width=\linewidth]{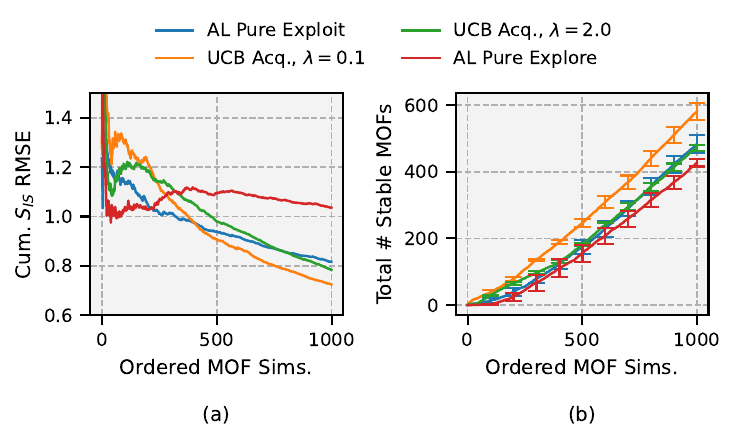}
    \caption{Explore/exploit trade-offs by MOF stability and total stable MOFs identified across AL strategies.
    (a) RMSE of stability for MOFs across explore/exploit workflows.
    (b) Stable MOFs identified across explore/exploit workflows.
    }
    \label{fig:mofa_explore_top_candidates}
\end{figure}

%% file: tables/utilization.tex
\begin{table}[t]
\caption{Hardware utilization across workflow steps.}
\label{tab:latency_steps}
\resizebox{\columnwidth}{!}{%
\begin{tabular}{cccc}
\toprule
    \textbf{Workflow Step} & \textbf{Alloc. Res.} & \textbf{\% Time Util.} & \textbf{\% Total CPU Util.} 
    \\
\midrule
    DiffLinker Fine-Tuning & 1 GPU & 80.7\% & - 
    \\
    DiffLinker Generation & 1 GPU & 16.1\% & - 
    \\
    AL Prioritization & 1 CPU & 28.9\% & 0.6\% 
    \\
    Assembly/Validation & 1 CPU & 88.6\% & 1.8\% 
    \\
    MD Simulation & 48 CPUs & 98.5\% & 94.6\% 
    \\
\bottomrule
\end{tabular}
}
\end{table}

%% file: sections/7_conclusions.tex
Generative AI models are rapidly being adopted in high-throughput screening workflows in areas like drug discovery and and MOF structure elucidation.
We have shown how our proposed queue prioritization algorithm can improve candidates produced by a generative AI model, increasing the number of suitable candidates identified by our screening workflow.
We demonstrate how acquisition functions can be tailored for multiple objectives to negotiate explore/exploit trade-offs.
We show how balancing exploration and exploitation objectives for queue prioritization ultimately leads to discovering a greater number of high-performing MOF materials.
In the future, we hope to demonstrate how AL can be used to prioritize candidates produced by generative AI for more complex, multi-faceted workflows.